\documentclass[runningheads]{llncs}
\usepackage{subfigure}
\usepackage{array,multirow,makecell}
\usepackage{multirow}
\usepackage{graphicx}
\usepackage{lscape}
\usepackage[dvipsnames]{xcolor}
\usepackage{booktabs,longtable,tabularx,caption,ragged2e,supertabular}

\usepackage{xurl}
\usepackage{hyperref}
\usepackage[doipre={DOI:~}]{uri} 
\usepackage{csquotes}
\usepackage{amsmath}

\usepackage{float}
\usepackage{lipsum}
\usepackage{multirow}
\usepackage{array}
\usepackage{amssymb}
\newcolumntype{L}[1]{>{\raggedright\let\newline\\\arraybackslash\hspace{0pt}}m{#1}}
\let\llncssubparagraph\subparagraph
\let\subparagraph\paragraph
\usepackage[compact]{titlesec}
\let\subparagraph\llncssubparagraph
\begin{document}
\title{FNDEX: Fake News and Doxxing Detection with Explainable AI}
%

%
\author{Dorsaf Sallami\inst{1,2}\orcidID{0000-0001-5077-3413} \and Esma Aïmeur\inst{1,3}\orcidID{0000-0001-7414-5454}}
\authorrunning{D. Sallami et al.}
%
\institute{Department of Computer Science and Operations Research (DIRO), University of Montreal, Canada\\ \and
\email{dorsaf.sallami@umontreal.ca}\\\and
\email{aimeur@iro.umontreal.ca}\\ }
\maketitle              
\begin{abstract}
The widespread and diverse online media platforms and other internet-driven communication technologies have presented significant challenges in defining the boundaries of freedom of expression. Consequently, the internet has been transformed into a potential cyber weapon. Within this evolving landscape, two particularly hazardous phenomena have emerged: fake news and doxxing. Although these threats have been subjects of extensive scholarly analysis, the crossroads where they intersect remain unexplored.
This research addresses this convergence by introducing a novel system. The Fake News and Doxxing Detection with Explainable Artificial Intelligence (FNDEX) system leverages the capabilities of three distinct transformer models to achieve high-performance detection for both fake news and doxxing. To enhance data security, a rigorous three-step anonymization process is employed, rooted in a pattern-based approach for anonymizing personally identifiable information. Finally, this research emphasizes the importance of generating coherent explanations for the outcomes produced by both detection models. Our experiments on realistic datasets demonstrate that our system significantly outperforms the existing baselines.
\end{abstract}

\section{Introduction}
\subsection{Background}
The internet has effectively turned the entire world into a global village, easily and quickly connecting people from different corners of the globe. However, it has also heightened various social harms, such as bullying and harassment, hate speech, disinformation, and radicalization. The exacerbation of these problems has wide-reaching impacts on individuals, communities, and society \cite{DigitalThreats}. Indeed, the online landscape has become a battleground where privacy, truth, and trust are continually tested. Within this volatile environment, the convergence of two troubling phenomena, doxxing and fake news, has raised serious concerns about the safety, security, and veracity of online communities.

Fake news, as a stand-alone menace, represents the deliberate spread of false or misleading information under the guise of credibility \cite{aimeur2023fake}. From conspiracy theories to fabricated narratives, fake news has the potential to influence public opinion \cite{tong2020fake} and sway elections \cite{calvillo2021individual}. On the other front, doxxing stands as a digital weapon wielded to expose personal and often sensitive information about individuals online \cite{karimi2022automated}. It can lead to real-world harm, causing victims to experience harassment, stalking, and even physical danger \cite{wang2019donttweetthis}. 
 
Each of them poses a potential risk and it is even more imperative to assess the hazardous implications arising from their convergence.
Both phenomena have real-world repercussions and can harm individuals and society as a whole. Moreover, these two phenomena can amplify each other's impact, making the consequences even more severe.
Doxxing combined with fake news yields a vicious circle of harm. False claims can be used to excuse and perpetuate doxxing, and exposing personal information can increase the impact of fake news. 

\subsection{Motivations: The Intersection of Doxxing and Fake News}
While fake news and doxxing are distinct, instances of intersection do exist. False information disseminated through fake news can contribute to the motivations behind doxxing. Conversely, doxxing may be employed as a form of retaliation against individuals associated with the creation or dissemination of fake news. Both phenomena give rise to substantial ethical and legal concerns, emphasizing the challenges of responsibly navigating the digital landscape.

\begin{figure}[h]
\centering
\includegraphics[width=\textwidth]{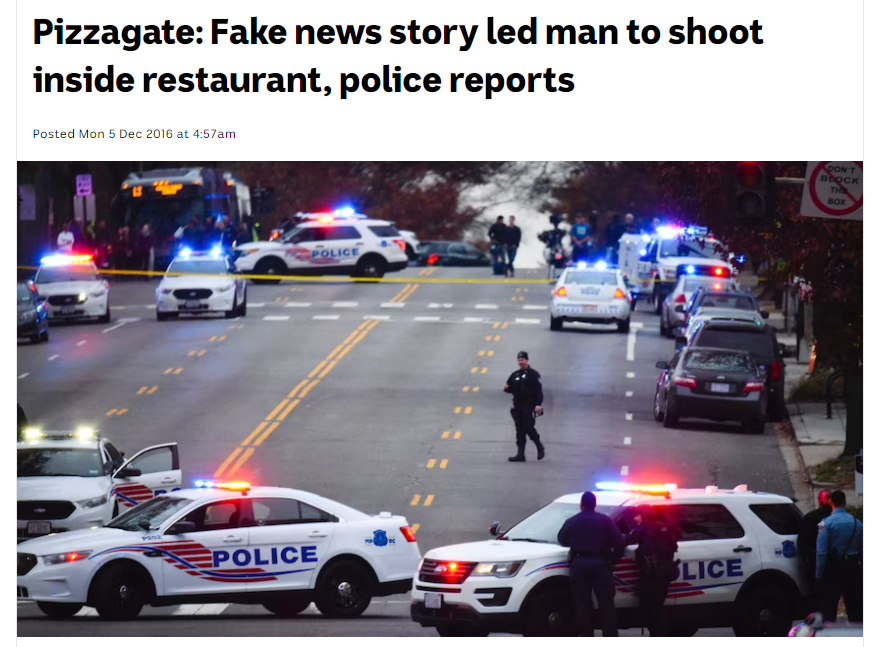}
\caption{Pizzagate conspiracy \cite{Pizzagate}. } \label{fig:mit}
\end{figure}

The correlation between fake news and doxxing warrants in-depth investigation. A notable example to highlight is the `Pizzagate' conspiracy theory (Figure \ref{fig:mit}). This conspiracy, which falsely implicated Hillary Clinton in a pedophilia ring, gained momentum through the spread of disinformation on social media, eventually leading to the exposure of personal information about innocent individuals (employees at the pizza parlor),  leading to a dangerous armed incident \cite{pohl2017cognitive}. This case underscores how the dissemination of fake news can contribute to doxxing, and potentially subjecting individuals to harm and harassment.

Furthermore, the convergence of COVID-19 vaccine disinformation and doxxing campaigns, which target vaccine researchers and healthcare professionals, has instilled a pervasive atmosphere of fear and reluctance to speak out within the healthcare sector. This correlation can be observed in the case of ``Plandemic" \cite{Plandemic}. This video disseminated misleading information about the COVID-19 pandemic and vaccines, ultimately culminating in the exposure of Dr. Anthony Fauci, a distinguished infectious disease expert, to doxxing. Dr. Fauci's private details were posted online, resulting in harassment and threats \cite{Plandemic}.

Fake news and doxxing often intertwine when false narratives are used to justify or legitimize the exposure of someone's private information. For instance, during a political campaign, fake news about a candidate's alleged misdeeds may be used to justify doxxing their personal details, aiming to harm their reputation or incite harassment. Fake news creators use private information to tailor their content to the beliefs, fears, and desires of target voters, making it more likely to resonate with them and deceive them \cite{PinfoinFakeNews}. By leveraging personal data, fake news purveyors can deceive the public by enabling the personalization of content to specific individuals or groups, increasing resonance, and reducing the likelihood of detection or opposition. Indeed, the Pew Research Center notes that false information about individuals or groups can lead to targeted harassment, discrimination, or other forms of abuse, which can compromise their privacy and security \cite{Report}.
In addition, targeted disinformation campaigns can involve the use of personal information to tailor the content and make it more convincing \cite{peter2019combating}. 
The intersection between fake news and doxxing is intricately interwoven with broader societal challenges, extending its reach into the realms of disinformation, online hate speech, and the erosion of democratic processes. 

\subsection{Contributions}
While there are separate studies dedicated to fake news and doxxing, to the best of our knowledge, there is currently no research that addresses both of these issues simultaneously. This research endeavor aims to delve into the complex realm of doxxing and fake news. Our objectives encompass not only the development of effective detection strategies for identifying these threats but also the implementation of anonymization techniques to safeguard individuals' privacy. Additionally, we propose an ethical and transparent pathway forward by incorporating eXplainable AI (XAI) methods to ensure that our solutions are both effective and accountable. Hence, our main contributions are: 
\begin{itemize}

    \item Developing an automated approach for detecting Doxxing using transformer models, diverging from traditional machine learning in Doxxing detection.

    \item Assessing three different transformer models to evaluate their effectiveness in detecting fake news.

    \item Conducting a three-step anonymization process using a pattern-based Personally Identifiable Information (PII) anonymization method to safeguard sensitive information in text data.

     \item Generating pertinent explanations for both detection models using explainable AI.
\end{itemize}

This paper proceeds as follows: Section 2 summarizes the related work. In Section 3, we delve into the details of the proposed approach. Section 4 presents the experimental implementation details. In Section 5, we explore and discuss the results. Finally, Section 6 concludes the paper with some discussion on the future directions.

\section{Literature Review}
The current study is grounded in two established research paradigms. Firstly, it encompasses methodologies aimed at automated fake news detection. Secondly, it encompasses the identification of doxxing and the exposure of private information. In this section, we will provide a concise overview of the pertinent literature in both of these domains.
\subsection{Fake News Detection}
Fake news has a long history with roots that predate even the invention of the printing press. While Google Trends Analysis\footnote{https://trends.google.com/trends/?geo=CA} reveals that the term "fake news" reached its peak in popularity during the 2016 US presidential election, its origins can be traced back to a time long before that \cite{burkhardt2017history}. Nevertheless, the concept of fake news remains a contentious issue, lacking a clear and universally accepted definition. The very definition of this concept and the way it is interpreted have increasingly become subjects of debate \cite{cunha2018fake,sallami2024deception}. The term "fake news" often encompasses disinformation, misinformation, and malinformation in public discourse, as highlighted by the Council of Europe \cite{wardle2017information}. Moreover, 
fake news may lead to a wide-ranging spectrum of harms and consequences, including malicious intents such as attempting to defraud individuals \cite{size2020publishing}, manipulate public opinion \cite{chambers2021truth}, or influence political processes \cite{van2020you}.

Fake news detection is an ever-expanding research topic that is gaining a lot of attention since there are still multiple challenges that need to be investigated. There are various research studies on fake news detection proposing different approaches. 
For instance, within the research community, various techniques have been employed to address this challenge. These include Natural Language Processing techniques \cite{sallami2023hype}, data mining approaches \cite{shu2020mining}, machine learning methods \cite{amri2021exmulf,hu2022deep}, recommender systems \cite{sallami2023trust}, social context-based techniques \cite{zhang2020overview}, and fake news propagation path \cite{raponi2022fake}.

Within the literature, several research endeavors have utilized diverse machine learning algorithms to combat the spread of false information. In the study \cite{mahir2019detecting}, the authors used different classifiers, including Support Vector Machine (SVM), Naïve Bayes, and Logistic Regression, to effectively identify fake news. Similarly, Aphiwongsophon et al.  \cite{aphiwongsophon2018detecting} contributed to the field by developing a technique specifically tailored for fake news detection on social media, leveraging both Naïve Bayes and SVM.
Furthermore, Ahmed et al. \cite{ahmed2017detection} employ a set of classifiers such as K-Nearest Neighbour (KNN), SVM, Logistic Regression, and Decision Tree in their efforts to identify and combat fake news. In parallel, Zhou et al. \cite{zhou2019network} proposed a network-based approach for uncovering patterns within false articles and fake news by harnessing classifiers, including SVM, KNN, Decision Tree, and Random Forest.

The use of deep learning algorithms has become increasingly prominent alongside traditional machine learning techniques. For instance, Kumar et al. \cite{kumar2020fake} crafted a model for identifying false news by employing Convolutional Neural Networks (CNN), Long Short-Term Memory (LSTM), and Bidirectional LSTM (Bi-LSTM). Similarly, Monti et al. \cite{monti2019fake} introduced a groundbreaking model grounded in geometric Deep Learning (DL), featuring a specialized CNN algorithm that extends the classical CNN framework.
Moreover, Girgis et al. \cite{girgis2018deep} developed a model that harnessed Recurrent Neural Networks (RNN) and CNN to dissect distinct facets of news within a sentence, incorporating LSTM, Gated Recurrent Unit (GRU), and CNN models. Additionally, Graph Neural Networks (GNN) is a distinctive approach for fake news detection, underscoring the ever-expanding application and versatility of deep learning algorithms within this critical domain \cite{wang2020fake,hu2019multi,qian2021knowledge}.

One prominently employed approach involves the use of transformers, as these algorithms have garnered substantial traction within the domain of natural language processing \cite{gillioz2020overview}.
In the realm of research, a spectrum of approaches has been employed. Some scholars, as demonstrated in the works of \cite{hande2021evaluating,mehta2021transformer,blackledge2021transforming}, have harnessed pre-trained transformers like BERT \cite{devlin2018bert}, capitalizing on their pre-training capabilities by fine-tuning them with fake news detection datasets.
Conversely, other researchers have delved into adapted transformers, as exemplified in the studies by \cite{rai2022fake,aggarwal2020classification}. These scholars have sought to improve existing transformer models for improved performance in fake news detection tasks. Their approach involves rigorous experimentation with various architectures and hyperparameters, evaluating their efficacy across diverse datasets.
Furthermore, domain-specific transformers have garnered attention and were fine-tuned on datasets tailored to specific domains such as politics, or health \cite{vijjali2020two,gundapu2021transformer}.

\subsection{Doxxing and Private Information Detection}
The term ``dox" has a somewhat ambiguous origin, but a widely accepted theory suggests it may have emerged as a condensed form of the word `` documents," particularly in the context of ``dropping documents". Its initial usage dates back to the 1990s \cite{douglas2016doxing}, when it came to signify the act of exposing someone's private and sensitive information online, often with the intention of embarrassing or intimidating them. 
Consequently, doxxing, also known as dropping doxx, is the act of releasing private, or personally identifiable information on the internet, often with malicious intentions \cite{anderson2021doxxing}.

Douglas \cite{douglas2016doxing} presents a comprehensive typology that delves into the multifaceted nature of doxxing, categorizing it into three distinct forms: deanonymizing doxxing, targeting doxxing, and delegitimizing doxxing. Within this framework, Douglas highlights the diverse losses or harm experienced by individuals subjected to these different forms of doxxing. Deanonymizing doxxing exposes the target's loss of anonymity, whether it pertains to their professional or personal life. In contrast, targeting doxxing results in the loss of obscurity as the doxxer typically reveals the individual's home address or other personal details online. Delegitimizing doxxing places the target's credibility in jeopardy, often by releasing evidence suggesting their involvement in deceptive or `immoral' activities. Douglas' typology provides a nuanced understanding of the varied consequences and dimensions of doxxing.

Several approaches have been proposed by different researchers for detecting sensitive information disclosures on X (formerly known as Twitter). Caliskan-Islam et al. \cite{caliskan2014privacy} introduced a method that scores users based on the disclosure of private information in their tweets, utilizing the detection of private information and subsequent calculation of privacy scores for users' timelines. Mehdy et al. \cite{mehdy2020user} developed a sentiment-aware privacy disclosure detection framework centered on neural networks, employing machine learning algorithms to identify sensitive information based on the sentiment expressed in the text. Confora et al.  \cite{canfora2018nlp} devised 97 heuristics rooted in recurrent patterns to uncover location and emotion information disclosures, relying on the identification of specific text patterns to detect sensitive information. In contrast, Deodhar et al. \cite{deodhar2017analysis} proposed an approach that not only identifies private information but also categorizes it. This approach includes the extraction of topics and features based on the dependency graph structures of tweets, as well as the consideration of location information and user mentions.

Other researchers have focused on detecting doxxing. For instance, researchers in \cite{karimi2022automated} propose a set of approaches for automatically detecting the disclosure of sensitive personal information, specifically doxxing, on X. It compares nine different approaches based on string-matching, heuristics, word embeddings, and contextualized string embeddings.
In a separate study \cite{snyder2017fifteen}, the authors address a significant knowledge gap surrounding doxxing by offering comprehensive, quantitative insights into this form of online harassment. The authors crafted and implemented a specialized tool capable of identifying dox files and quantifying various aspects of doxxing, including its prevalence, content, targets, and the consequences observed on well-known dox-posting platforms.

The intersection between fake news and doxxing forms a deeply concerning and multifaceted issue, posing a significant threat to both the integrity of online information ecosystems and the personal safety and privacy of individuals. 
While the existing body of literature has separately addressed both fake news and doxxing threats, a comprehensive approach that combines these two distinct issues has not been previously explored. Therefore, the present study represents pioneering research in integrating these two domains, as far as our knowledge extends. Moreover, it is noteworthy that prior methodologies for doxxing detection primarily relied on conventional machine learning techniques. In contrast, this study leverages transformer models, which have garnered significant attention for their exceptional capabilities, signifying a notable shift towards more advanced and effective methods in addressing this challenge.

\section{Proposed Methodology}
\subsection{Problem Statement}
Given a dataset $N$ consisting of news posts, where $N = \{n_1, n_2, \ldots, n_n\}$, with each news post $n_x$ containing text content $t_x$ and an authenticity label $p_x$ (where $p_x = 0$ for real news and $p_x = 1$ for fake news), we aim to develop functions and strategies to address the challenges of fake news and doxxing detection and prevention.
Specifically, we intend to design a fake news detection function $F$ that maps the input text $t_x$ to the predicted authenticity label: $p_x = F(t_x)$. Additionally, we introduce a doxxing detection function $D$ to identify doxxing within the text content: $d_x = D(t_x)$.

For prevention, we introduce a text anonymization process denoted by the function $A$. Given the input text $t_x$, the anonymization function $A$ transforms it to a protected text $\hat{t}_x$:
$\hat{t}_x = A(t_x)$.
The aim of text anonymization is to safeguard sensitive information within the text content while preserving the overall message, reducing the risk of privacy breaches and harm associated with doxxing. 

To ensure transparency and interpretability in our methodology, we incorporate XAI techniques. 
For the fake news detection model, denoted as \(F\), we introduce an explanation function \(E_F\) that takes the input text \(t_x\) and produces an explanation \(e_{F_x}\) for the authenticity prediction \(p_x\): $e_{F_x} = E_F(t_x)$.
Similarly, for the doxxing detection model \(D\), we employ another explanation function \(E_D\) to generate an explanation \(e_{D_x}\) for the doxxing prediction \(d_x\): $e_{D_x} = E_D(t_x)$.
These explanations \(e_{F_x}\) and \(e_{D_x}\) provide insights into how the models arrive at their predictions, ensuring transparency and helping users understand the decision rationale. 
Table \ref{tab:annotation} offers a detailed overview of the notation, clarifying the key parameters for the explanation framework.

\begin{table}[h]
\centering
\caption{Terminology and Notations.}\label{tab:annotation}
\begin{tabular}{cl}
\hline
\textbf{Notation} & \textbf{Description}                             \\ \hline
$N$                                & The dataset of news posts, where $N = \{n_1, n_2, \ldots, n_n\}$. \\ \hline
$n_x$                              & Each news post in the dataset.                                    \\ \hline
$t_x$                              & The text content of each news post $n_x$.                         \\ \hline
$p_x$                              & The authenticity label for each news post.                        \\ \hline
$d_x$                              & The doxxing label for each news post.                        \\ \hline
$\hat{t}_x$                              & The anonymized text.             \\ \hline
$F$                                & The fake news detection function.                                 \\ \hline
$D$                                & The doxxing detection function.                                    \\ \hline
$A$                                & The text anonymization function.                                  \\ \hline
$E_F$                              & The explanation function for the fake news detection model $F$.   \\ \hline
$E_D$                              & The explanation function for the doxxing detection model $D$.      \\ \hline
\end{tabular}
\end{table}

\subsection{Approach overview}
\begin{figure}[h]
\centering
\includegraphics[width=0.8\textwidth]{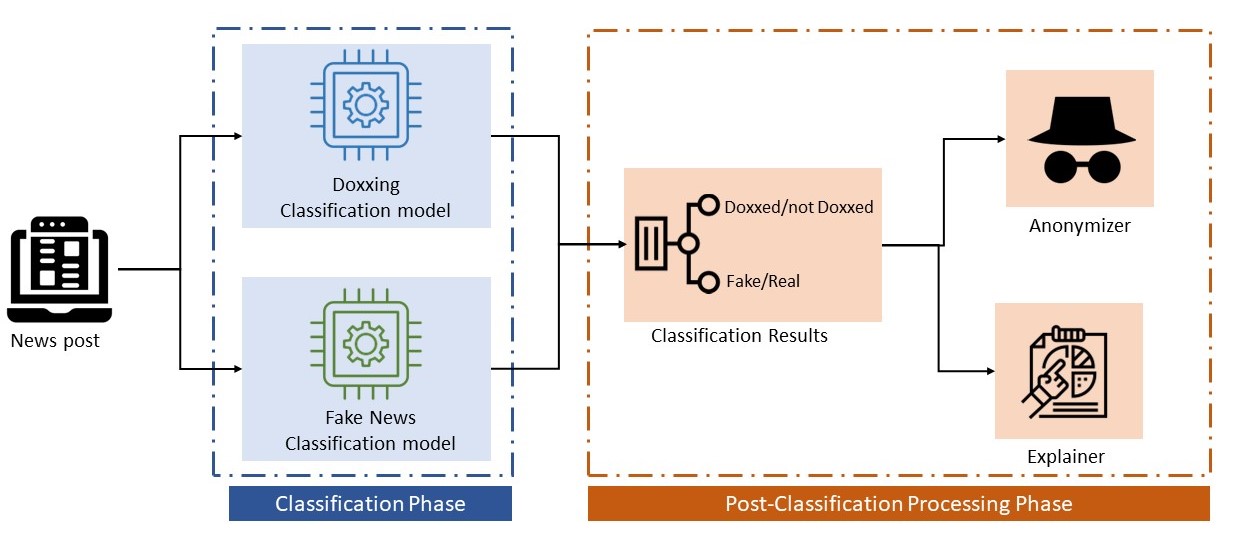}
\caption{An illustration of the architecture of the Fake News and Doxxing detection with EXplainable AI (FNDEX) system.} \label{fig:method}
\end{figure}

In Figure \ref{fig:method}, our proposed framework is depicted, comprising primarily of two phases: the classification phase and the post classification processing phase. The initial step involves the reception of news posts as input, at which point two pre-defined classifiers are applied—one for detecting fake news and another for identifying doxxing. Subsequently, in the ensuing phase, the outcomes of these two modes are presented, indicating whether the news is genuine or not, and whether it has been doxxed or not. If the news is deemed to be doxxed, an anonymization process is employed to safeguard against the disclosure of private information. Finally, the explainer component offers comprehensive explanations for both classifiers.
\subsection{Classification phase}
\begin{figure}[h]
\centering
\includegraphics[width=0.8\textwidth]{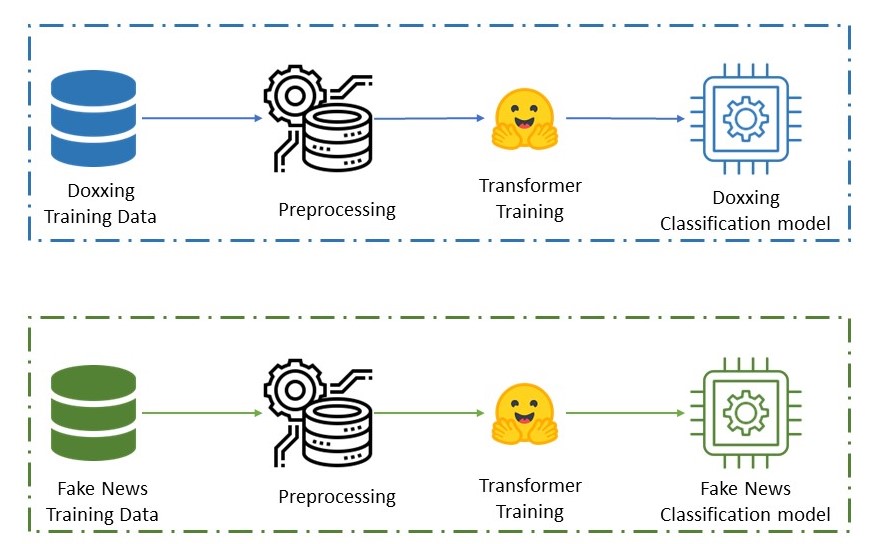}
\caption{Offline Phase: Models Training} \label{fig:off}
\end{figure}

The initial phase involves the application of two classifiers designed for the identification of fake news and doxxing. Both of these models undergo offline training. Figure \ref{fig:off} provides a zoomed-in view of the training process.
Starting with separate datasets for each task, specifically for fake news detection and doxxing identification, an identical process is systematically executed. This process encompasses preprocessing and transformer training, ultimately yielding a distinct classifier for each task.
\subsubsection{Preprocessing}\mbox{}\\ To ensure the dataset is suitably preprocessed for input into our models, we applied standard data processing techniques. This encompassed employing Huggingface's AutoTokenizer \cite{AutoClasses} to tokenize the text as per our specific requirements, performing lemmatization, and removing both stop words and punctuation. Furthermore, we conducted label encoding to effectively convert categorical data into a numerical format.

\subsubsection{Transformer training}\mbox{}\\
Our research builds upon the use of transformers.

\textbf{Transformer architecture:} The Transformer architecture, as described in \cite{gillioz2020overview}, employs an encoder-decoder structure. It takes an input sequence denoted as $X = (x_1, \ldots, x_N)$ and generates a corresponding latent representation referred to as $Z = (z_1, \ldots, z_N)$. Notably, owing to the autoregressive nature of this model, the output sequence $Y_M = (y_1, \ldots, y_M)$ is produced incrementally, with each element, such as $y_M$, relying on both the latent representation $Z$ and the previously generated sequence $Y_{M-1} = (y_1, \ldots, y_{M-1})$ for its creation.

Both the Encoder and the Decoder components utilize an identical Multi-Head Attention layer. This specific Attention layer functions by mapping a query denoted as $Q$ and a set of keys represented as $K$ to a weighted sum of corresponding values denoted as $V$. It is important to note that, for technical considerations, a scaling factor of $\sqrt{\frac{1}{d_k}}$ is applied in the computation \cite{gillioz2020overview}, as mentioned in the following equation:
\begin{equation*}
\text{Attention}(Q,K,V) = \text{Softmax} \left( \frac{QK^T}{\sqrt{d_k}} \right) V
\end{equation*}
\textbf{Transformers used:} In our experimentation, we explored the capabilities of three different transformer models. First, we employed \textit{BERT} (Bidirectional Encoder Representations from Transformers) \cite{devlin2018bert}, a language representation model that utilizes bidirectional pre-training by considering both the left and right contexts of unlabeled text. BERT's pre-training process involves two primary objectives: Masked Language Modeling (MLM), where 15\% of the words in a sentence are randomly masked and the model predicts these masked words, and Next Sentence Prediction (NSP), where the model assesses whether two concatenated masked sentences follow each other. Fine-tuning BERT for specific tasks requires just one additional output layer.
Next, we explored \textit{DistilBERT} \cite{sanh2019distilbert}, a model designed to create a more compact version of the BERT model. DistilBERT achieves a 40\% reduction in size compared to BERT while retaining 97\% of its language understanding capabilities and demonstrating a 60\% faster processing speed. This distillation process makes it an efficient choice for various natural language processing tasks.
Lastly, we delved into \textit{RoBERTa} \cite{liu2019roberta}, a model introduced as an enhancement of BERT through an extensive replication study. RoBERTa's improvements include longer training durations, larger batches, elimination of the next sentence prediction objective, training on longer sequences, and dynamic adjustments to the masking pattern used in training data. These modifications collectively contribute to RoBERTa's superior performance in various language understanding tasks.
\subsection{Post classification processing phase}
As previously stated, following the classification phase, the outcomes will be presented to the user. Additionally, in the event that the text is categorized as containing personally identifiable information (PII), the anonymization process will be initiated to protect sensitive data. Subsequently, the explainer module will provide explanations.
\subsubsection{Anonymizer}\mbox{}\\
The anonymizer is based on a pattern-based PII anonymization method to protect sensitive information in text data. It involves three different steps:
\begin{enumerate}
    \item \textbf{Pattern Identification:} The first step in this anonymization technique is to identify specific patterns or regular expressions that correspond to different types of PII. These patterns are designed to match the structure of PII, such as a name's format, an email address format, or a social security number format.

\item \textbf{Anonymization Placeholders:} Following this identification step, we establish anonymization placeholders, which are used to replace the identified PII patterns in the text. These placeholders do not reveal any specific individual's identity but retain the format and structure of the original data.

\item \textbf{Pattern Replacement:} Lastly, we replace PII patterns in the text with their corresponding anonymization placeholders. This process is accomplished through the use of regex pattern matching and replacement functions.
\end{enumerate}

Table \ref{tab:anonym} presents an exhaustive mapping that aligns personally identifiable information Patterns with their respective Regular Expressions and the associated Anonymization Placeholders. 

\begin{landscape}
\begin{table}[h]
\centering
\caption{Mapping of Personally Identifiable Information (PII) Patterns to Regular Expressions and Anonymization Placeholders.}\label{tab:anonym}
\begin{tabular}{lll}
\hline
\textbf{PII patterns} & \textbf{Regular expressions}                                                                                                                                                                                           &\textbf{Placeholders}                      \\ \hline
Name         & \textbackslash{}b{[}A-Z{]}{[}a-z{]}+\textbackslash{}s{[}A-Z{]}{[}a-z{]}+\textbackslash{}b                                                                                                                     & FULL\_NAME                       \\ \hline
Email        & \textbackslash{}b{[}A-Za-z0-9.\_\%+-{]}+@{[}A-Za-z0-9.-{]}+\textbackslash{}.{[}A-Za-z{]}\{2,\}\textbackslash{}b                                                                                               & EMAIL\_ADDRESS                   \\ \hline
Phone        & \textbackslash{}b(?:\textbackslash{}+\textbackslash{}d\{1,2\}\textbackslash{}s?)?\(?\d{3}\)?{[}-.\textbackslash{}s{]}?\textbackslash{}d\{3\}{[}-.\textbackslash{}s{]}?\textbackslash{}d\{4\}\textbackslash{}b & PHONE\_NUMBER                    \\ \hline
SSN          & \textbackslash{}b\textbackslash{}d\{3\}{[}-.\textbackslash{}s{]}?\textbackslash{}d\{2\}{[}-.\textbackslash{}s{]}?\textbackslash{}d\{4\}\textbackslash{}b                                                      & SSN                              \\ \hline

Address       & \textbackslash{}d+\textbackslash{}s{[}A-Za-z\textbackslash{}s{]}+\textbackslash{}s(?:Street|Ave|St|Dr|Rd|Blvd|Ln|Ct)\textbackslash{}b                                                                         & ADDRESS                          \\ \hline
CCN          & \textbackslash{}b(?:\textbackslash{}d\{4\}{[}-.\textbackslash{}s{]}?)\{3\}\textbackslash{}d\{4\}\textbackslash{}b                                                                                             & CREDIT\_CARD\_NUMBER             \\ \hline
DL           & \textbackslash{}b{[}A-Z0-9{]}\{8,10\}\textbackslash{}b                                                                                                                                                        & DRIVER\_LICENSE                  \\ \hline

Passport    & \textbackslash{}b{[}A-Z0-9{]}+\textbackslash{}b                                                                                                                                                               & PASSPORT\_NUMBER                 \\ \hline

IP           & \textbackslash{}b(?:\textbackslash{}d\{1,3\}\textbackslash{}.)\{3\}\textbackslash{}d\{1,3\}\textbackslash{}b                                                                                                  & IP\_ADDRESS                      \\ \hline

VIN          & \textbackslash{}b{[}A-HJ-NPR-Z0-9{]}\{17\}\textbackslash{}b                                                                                                                                                   & VEHICLE\_IDENTIFICATION\_NUMBER  \\ \hline
Social Media & @\textbackslash{}w+                                                                                                                                                                                           & SOCIAL\_MEDIA\_USERNAME          \\ \hline

\end{tabular}
\end{table}
\end{landscape}

\subsubsection{Explainer}\mbox{}\\
In the realm of artificial intelligence applications, establishing trust is paramount to facilitate informed decision-making, as a lack of trust can lead to the disregard of AI-generated advice \cite{shin2021effects}. Here, Explainable AI methods play a pivotal role by unraveling the complex mathematical underpinnings behind model-generated predictions \cite{guo2020explainable}. These explainable models offer users coherent and meaningful explanations, ultimately aiming to foster confidence in the system's outcomes. This enhanced understanding not only raises awareness of the dangers associated with such content but also influences users' future behavior. 
Recognizing the significance of transparency and interpretability in AI systems, we ensure that both our fake news detection model and doxxing detection model are explainable, enabling users to understand the decision-making processes behind these models.

In our methodology, we employ LIME (Local Interpretable Model-Agnostic Explanations) \cite{ribeiro2016should}, an algorithm designed to faithfully elucidate the predictions of any classifier or regressor. It accomplishes this by creating a localized approximation of the model in question using an interpretable model.
One of the standout features of LIME is its remarkable accessibility and simplicity. LIME's model-agnostic nature allows it to seamlessly integrate with virtually any machine learning model. It accomplishes this by treating the model as a distinct black-box entity and generating explanations for it. Furthermore, LIME goes beyond providing mere model-level explanations, offering insightful, granular explanations for each individual observation. Additionally, LIME's interpretability shines through in its capacity to furnish explanations that are grounded in the input features, avoiding the use of abstract or obscure attributes.

The rationale for this framework is deeply rooted in the urgency of addressing the combined threats of fake news and doxxing. With our methodology, we not only protect online safety and information integrity but also fortify the foundations of a thriving digital society where accurate information, privacy, and security are cherished and upheld.

\section{Experimental Implementation}
\subsection{Datasets}
To evaluate the performance of the fake news detection task, we conducted an experiment on the Kaggle dataset \cite{ahmed2018detecting}. This dataset is in the English language and comprises two main subsets: the training set and the test set. The training set contains \textit{35,918} data samples, while the test set has \textit{8,980} data samples. These samples are news articles, and they are categorized into two classes: Fake and Real news. Within the dataset, there are \textit{23,481} articles classified as fake news and \textit{21,417} articles classified as real news. This dataset is ideal for natural language processing tasks, especially for classifying text to differentiate between fake and real news articles.

In the domain of Doxxing detection, a significant challenge we faced revolved around the absence of an openly accessible dataset. This scarcity can be ascribed to the paramount need to protect the privacy of individuals who engage with online platforms. Recognizing the critical importance of this matter, we took proactive steps to tackle this issue. As part of this endeavor, we established communication with the authors of a pertinent scholarly publication referenced as \cite{karimi2022automated}, and they provided the tweet IDs. 
Our approach in utilizing this dataset closely mirrored the methodology outlined in the original research paper. However, it is important to acknowledge that the current API is limited. In the course of our data collection, we encountered instances where tweets had been deleted/not found. As a result, the final dataset comprises \textit{1,456} instances of doxxed content and \textit{863} instances of non-doxxed content. 
\subsection{Experimental Setup}
The experimental setup involved utilizing Google Colab Pro and PyTorch to conduct the experiments. For the pre-trained models, we used the Hugging Face library\footnote{https://huggingface.co/docs/transformers/index}. 
The specific hyperparameters used can be found in Table \ref{tab:expVal}.

\begin{table}[h]
\centering
\caption{Hyperparameters used for Training.}
\label{tab:expVal}
\begin{tabular}{ll}
\toprule
\textbf{Hyperparameters }                & \textbf{Experimental value} \\ \midrule
Number of epochs & 5                 \\ 
Batch size              & 8                 \\ 
Warmup steps             & 500               \\
Weight decay              & 0.01              \\ 
Logging steps           & 400               \\ 
\bottomrule
\end{tabular}
\end{table}
\subsection{Evaluation Metrics}
The performance of the algorithms for each task is measured in terms of detection accuracy, precision, recall, F1-score, which are calculated using the equations (1)–(4), respectively.

\begin{equation}
\text{Detection accuracy} = \frac{TN + TP}{TN + TP + FN + FP} \times 100 \tag{1}
\end{equation}

\begin{equation}
\text{Precision} = \frac{TP}{TP + FP} \times 100 \tag{2}
\end{equation}

\begin{equation}
\text{Recall} = \frac{TP}{TP + FN} \times 100 \tag{3}
\end{equation}

\begin{equation}
\text{F1-Score} = \frac{2 \times \text{Precision} \times \text{Recall}}{\text{Precision} + \text{Recall}} \times 100 \tag{4}
\end{equation}

where TP is true positive, TN is true negative, FN is false positive, and FN is false negative.

\subsection{Baselines}
To assess the performance of the two models, we compared them against established baselines:
\begin{itemize}
    \item  The different baselines for fake news detection are: (1) Triple BERT \cite{mehta2021transformer}, which employs multiple BERT models with shared weights to effectively handle a range of inputs and leverage additional contextual information, including justifications and metadata. (2) FNR \cite{ghorbanpour2023fnr} relies on similarity and transformer-based methods to detect fake news, utilizing both textual and image content from news sources. (3) SENAD \cite{uppada2022novel} introduces an authenticity scoring mechanism and incorporates user-centric metrics such as Following-followers ratio, account age, bias, and more to evaluate the credibility of user engagement with news articles.

\item The baselines for doxxing detection are: (1) In \cite{karimi2022automated}, a stratified 10-fold cross-validation method is used to ensure an equitable representation of both classes in the training and testing datasets, preventing overfitting or bias in the Support Vector Machines classifier. (2)  \cite{canfora2018nlp} detects sensitive information in social network posts by identifying recurring patterns in natural language frequently used by users to disclose private details. (3)  \cite{deodhar2017analysis} incorporates the Stanford Dependency Parser for dependency parsing, which captures grammatical relations between words in sentences. The Chi-square method is used to select informative features while eliminating irrelevant and redundant attributes from the data.
\end{itemize}

\section{Results and Discussion }
\subsection{Assessing Performance}
Table \ref{tab:results} offers a comprehensive breakdown of the performance metrics for each model across various datasets.

In the Doxxing Detection task, the comparison between baseline models and transformer-based models reveals a striking disparity in performance. The best-performing baseline model achieved respectable results with an accuracy of 96.61\%, precision of 97.74\%, recall of 97.28\%, and an F1 score of 97.51\%. However, the transformer-based models, including BERT, DistilBERT, and RoBERTa, exhibited remarkable superiority. RoBERTa, in particular, stood out with an astonishing accuracy of 99.99\%, precision of 100\%, recall of 100\%, and an F1 score of 99.98\%. These results clearly demonstrate that transformer-based models outclass the baseline models, providing significantly higher accuracy and precision in the challenging task of Doxxing Detection.

In the realm of Fake News Detection, a similar pattern emerges when comparing baseline models to transformer-based models. The best-performing baseline model achieved a reasonably strong performance, boasting an accuracy of 93.7\%, precision of 92.6\%, recall of 95\%, and an F1 score of 93.7\%. However, the transformer-based models, including BERT, DistilBERT, and RoBERTa, once again demonstrated their prowess. For instance, RoBERTa achieved an impressive accuracy of 99.65\%, precision of 99.89\%, recall of 99.44\%, and an F1 score of 99.66\%. These results highlight the clear advantage of transformer-based models over baseline models in fake news detection, as they consistently deliver higher accuracy and precision.
\begin{table}[h]
\centering
\caption{Evaluation Results.}
\label{tab:results}
\begin{tabular}{lllllll}
\hline
\textbf{Task}                               & \multicolumn{2}{l}{\textbf{Model}}               & \textbf{Accuracy} & \textbf{Precision} & \textbf{Recall} & \textbf{F1 score} \\ \hline
\multirow{6}{*}{\begin{tabular}[c]{@{}l@{}}Doxxing\\ Detection\end{tabular}} & \multirow{3}{*}{Baselines} &     Younes \textit{et al.} \cite{karimi2022automated}       &   96.61       &    97.74       &   97.28     &      97.51    \\ \cline{3-7} 
                                   &                            &         Canfora \textit{et al.}  \cite{canfora2018nlp} & 90.8         &  93.6         &  69.8      & 80         \\ \cline{3-7} 
                                   &                            &      Deodhar \textit{et al.} \cite{deodhar2017analysis}    &    92      & 93          & 92       &  92        \\ \cline{2-7} 
 &
  \multirow{3}{*}{\begin{tabular}[c]{@{}l@{}}Transformers\\ used\end{tabular}} &
  BERT &
  99.97 &
  99.95 &
  100 &
  99.97 \\ \cline{3-7} 
                                   &                            & DistilBERT & 99.97    & 99.95     & 100    & 99.97    \\ \cline{3-7} 
                                   &                            & RoBERTa    & 99.99    & 100       & 100    & 99.98    \\ \hline
\multirow{6}{*}{\begin{tabular}[c]{@{}l@{}}Fake news\\ Detection\end{tabular}} &
  \multirow{3}{*}{Baselines} &
Mehta \textit{et al.} \cite{mehta2021transformer}  & 74
   &69.6
   &85.4
   &76.8
   \\ \cline{3-7} 
                                   &                            &      Ghorbanpour  \textit{et al.}  \cite{ghorbanpour2023fnr}  & 78.9         &   78        &      85  &     82     \\ \cline{3-7} 
                                   &                            &        Uppada  \textit{et al.}\cite{uppada2022novel}  & 93.7         & 92.6          &  95      &   93.7       \\ \cline{2-7} 
 &
  \multirow{3}{*}{\begin{tabular}[c]{@{}l@{}}Transformers\\ used\end{tabular}} &
  BERT & 99.24    & 99.08     & 99.71  & 99.17 \\ \cline{3-7} 
                                   &                            & DistilBERT & 99.01    & 98.14     & 100    & 99.06    \\ \cline{3-7} 
                                   &                            & RoBERTa    &
  99.65 &
  99.89 &
  99.44 &
  99.66    \\ \hline
\end{tabular}%

\end{table}
\subsection{Interpreting Transformers' Attention}
\begin{figure}[h]
\centering
\includegraphics[width=0.8\textwidth]{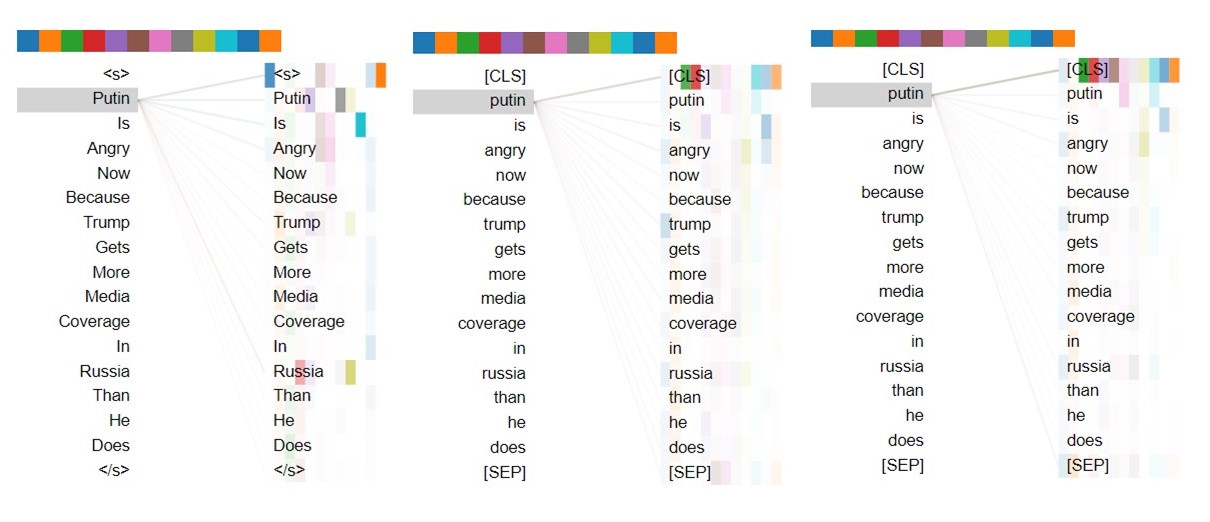}
\caption{RoBERTa BERT (left), BERT (center), and DistilBERT (right) attention to the word \textit{Putin}. } \label{fig:inter}
\end{figure}

In order to delve deeper into the performance of the three transformer models, we undertook a comprehensive visual analysis of attention weights in pre-trained models with the aid of the BertViz library \cite{vig2019bertviz}. This interactive tool is designed for visualizing attention in Transformer language models, offering multiple perspectives that provide unique insights into the attention mechanism. Figure \ref{fig:inter} depicts the head view on a tweet: "Putin Is Angry Now Because Trump Gets More Media Coverage In Russia Than He Does".
This figure illustrates why the RoBERTa model performed slightly better than other models. For instance, BERT and DistilBERT associate the token `Putin' primarily with the token `is'. In contrast, the RoBERTa model not only associates `Putin' with `is' but also links it with `Russia'. As a result, the RoBERTa model more effectively captures the context.

By visualizing attention, our aim is to leverage interpretability to understand why RoBERTa performs differently from BERT and DistilBERT. By making the decision-making processes of these models transparent, we can discern the reasons behind their varying outcomes, thereby shedding light on their unique behaviors.

\subsection{Privacy Utility Trade-Off}
In this section, we present the results obtained through the Anonymizer. Table \ref{tab:exempleAnonym} showcases illustrative examples that demonstrate the transformation of the original text into an anonymized version after applying the Anonymizer. It is important to note that these examples are synthetic and are not representative of real-world data, they are solely generated for the purpose of demonstrating the anonymization process.

\begin{table}[h]
\centering
\caption{Anonymized Text Examples.}\label{tab:exempleAnonym}
\begin{tabular}{ll}
\hline
\textbf{Original Text}                                                                                                                                                                               & \textbf{Anonymized Text}                                                                                                                                                                                                     \\ \hline
\begin{tabular}[c]{@{}l@{}}I met John Smith at john@example.com \\ and called (123) 456-7890. \\ My SSN is 123-45-6789\\ My VIN is 1HGCM82633A400000, \\ and my X username is @user123.\end{tabular} & \begin{tabular}[c]{@{}l@{}}I met FULL\_NAME at EMAIL\_ADDRESS \\ and called PHONE\_NUMBER. \\ My SSN is SSN , My VIN is \\ VEHICLE\_IDENTIFICATION\_NUMBER, \\ and my X username is \\ SOCIAL\_MEDIA\_USERNAME.\end{tabular} \\ \hline
\begin{tabular}[c]{@{}l@{}}Mary Johnson's email is  mary@example.com,\\ and her phone number \\ is (987) 654-3210. She lives at 456 Elm St.\end{tabular}                                             & \begin{tabular}[c]{@{}l@{}}FULL\_NAME's email is EMAIL\_ADDRESS, \\ and her phone number is PHONE\_NUMBER. \\ They live at ADDRESS\end{tabular}                                                                              \\ \hline
\end{tabular}
\end{table}
Anonymization is primarily carried out to protect the privacy of individuals whose data is being processed or shared. While preserving privacy is crucial, data also needs to maintain its utility and usefulness. Therefore, we conducted an assessment of anonymization effectiveness, as illustrated in Table \ref{tab:effecAnonym}.

\begin{table}[h]
\centering
\caption{Anonymization Performance Metrics.}\label{tab:effecAnonym}
\begin{tabular}{ll}
\hline
\textbf{Metric}              & \textbf{Value} \\ \hline
Cosine Similarity   & 0.849 \\ \hline
Jaccard Similarity  & 0.844 \\ \hline
Semantic Similarity & 0.950 \\ \hline
BLUE                & 0.84  \\ \hline
Normalized Certainty Penalty (NCP)                & 0.006 \\ \hline

\end{tabular}
\end{table}

The cosine similarity score of 0.849, which is close to 1, signifies a notable degree of similarity between the original and anonymized texts. This implies that the anonymization process has had minimal impact on the texts, resulting in a high degree of similarity between the anonymized and original versions. The Jaccard Similarity score of 0.844 between the two texts indicates a relatively high level of similarity in terms of the words or tokens utilized in both texts. Additionally, a semantic similarity score of 0.950 suggests an exceptionally strong resemblance between the items under comparison. Moreover, the BLEU score \cite{papineni2002bleu} of 0.84 implies that the anonymized text closely mirrors the original text, affirming the successful preservation of the text's overall meaning and structure during the anonymization process. Finally, we employed the Normalized Certainty Penalty (NCP) data utility metric \cite{terrovitis2008privacy}, and the recorded NCP value of 0.006 demonstrates that the anonymization process incurred only minimal information loss.

Based on these findings, it is evident that placeholders are an effective means of eliminating sensitive information, thereby enhancing the safeguarding of privacy and the protection of sensitive data. Notably, placeholders exhibit proficiency in retaining the contextual integrity of the text, which is a pivotal consideration. The adoption of placeholders expertly strikes a harmonious balance between the imperatives of privacy protection and data utility. This balance represents the essence of the Privacy Utility Trade-Off in the context of text anonymization.

In the Privacy Utility Trade-Off, privacy concerns are inherently pitted against the need for data utility. On one hand, we must ensure the anonymity of sensitive information to protect individuals' privacy and comply with data protection regulations. On the other hand, we need the data to remain useful for analysis. Indeed, placeholders excel in this regard by allowing the removal or substitution of sensitive data without compromising the overall structure and meaning of the text. This preservation of contextual integrity ensures that the anonymized data can still be utilized effectively for various applications. Thus, the adoption of placeholders represents a significant step towards achieving a balance between privacy protection and data utility in the realm of text anonymization.

\subsection{Models Interpretability Through LIME }
For the explainer phase, we employ the LIME Text Explainer \cite{ribeiro2016should}. We introduce an independent textual instance to the interpreter. Multiple iterations of the original text are generated, wherein a specified number of randomly selected words are systematically omitted. This newly generated synthetic data is subsequently categorized into distinct groups, differentiating between ``fake" and ``real" or ``doxxed" and ``not doxxed". Consequently, by observing the impact of specific keywords' presence or absence, we can assess their influence on the classification of the designated text. 

\begin{figure}[h]
\centering
\includegraphics[width=\textwidth]{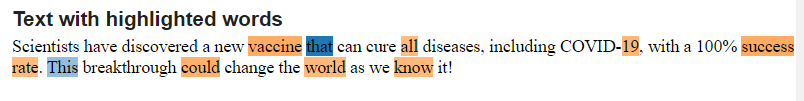}
\caption{LIME explanations for fake news detection.} \label{fig:lime1}
\end{figure}

\begin{figure}[h]
\centering
\includegraphics[width=\textwidth]{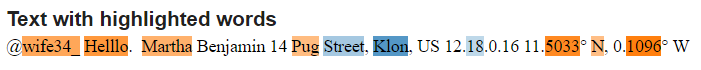}
\caption{LIME explanations for Doxxing detection.} \label{fig:lime2}
\end{figure}

The LIME output comprises a set of explanations delineating the individual feature's influence on the prediction of a given data sample. The resultant value provides a concise representation of each word's contribution to the classification of the text instance into a particular category. The LIME interpretations for both the fake news and doxxing detection tasks are illustrated in Figures \ref{fig:lime1} and \ref{fig:lime2} respectively. In these figures, the use of the color orange represents the fake or doxxed class, while the color blue signifies the real or not doxxed class. These visual representations
facilitate our comprehension of which words within the text exert the most significant influence on the model's ultimate prediction.

As an illustration, in the doxxing detection model, we observe that words highlighted in orange (representing the doxxed class) include social media usernames, names, and addresses. This observation implies that the model places considerable reliance on these particular terms to discern whether the text contains doxxing information.

In an age where trust in online information is precarious, transparency becomes the linchpin of credibility. By providing clear and detailed explanations for the classifications made by our detectors, users can assess the authenticity of the content they encounter, fostering a culture of informed media consumption. This transparency reinforces trust not only in the framework itself but also in the broader online ecosystem.
\subsection{Impact on Online Safety}
The proposed framework offers a holistic solution that holds the potential to enhance online safety by addressing the intersection of fake news and doxxing. By proactively detecting and responding to these dual threats, the framework aims to mitigate the risks associated with online harassment, privacy violations, and the spread of misinformation, thus creating a safer and more secure digital environment. It contributes to a significant reduction in online harassment, safeguarding individuals from the emotional distress and fear that often accompanies such incidents. By protecting privacy through the identification and prevention of doxxing attempts, it empowers users to navigate online spaces with a greater sense of personal security. Additionally, by actively countering misinformation, the framework helps preserve the integrity of information ecosystems, fostering a culture of trust and reliability. The educational aspect, through the explainer component, further empowers users with the knowledge to critically assess the information they encounter, while enhancing trust in online spaces. In sum, the framework is a comprehensive and proactive approach that equips individuals to engage in online communities with greater confidence, security, and trust in the accuracy of the information they encounter.

\subsection{Ethical Considerations}
Ethical considerations are at the heart of our framework, and we recognize the importance of transparency in addressing the specific ethical concerns our study addresses. The anonymization component is inherently designed to safeguard individuals' privacy, a central ethical concern in our research. Our foremost ethical concern is the preservation of individuals' privacy. We acknowledge that the data we collected and analyzed can be sensitive, and our primary goal is to prevent any potential harm, especially in the case of doxxing targets included in our research. By ensuring the privacy of these individuals, we are adhering to ethical principles related to data protection and individual rights. Moreover, to maintain ethical integrity, we meticulously adhered to a rigorous data collection policy. This policy mandates the exclusive use of publicly available information, thereby respecting the privacy boundaries of individuals and complying with data access restrictions. As an additional ethical precaution, we refrained from presenting the exact content of tweets or revealing any personal data contained within our dataset. This decision aligns with ethical guidelines related to non-disclosure of personal information without consent and minimizes the risk of unintended exposure. Our meticulous approach underscores our unwavering commitment to mitigating any possible harm that may result from our research. This includes potential harm to the individuals involved as well as broader societal implications. By taking these precautions, we aim to uphold the highest ethical standards throughout the study.

Our study addresses specific ethical concerns related to privacy preservation, data collection, avoidance of inadvertent exposure, and harm mitigation. We are dedicated to upholding these ethical principles as we conduct our research, ensuring that our work aligns with the values of responsible and ethical research practices.

\section{Conclusion and Future work}
While free speech is a fundamental right, the use of false information or the exposure of private data often has legal consequences, highlighting the need for effective detection and enforcement mechanisms.
Within the scope of this paper, we present fake news and doxxing detection with an explainable AI approach. FNDEX is a novel system designed for the dual purpose of detecting fake news and identifying doxxing instances within online text. This system not only anonymizes sensitive data but also offers detailed explanations to users, explaining the reasoning behind system decisions.
Our proposed framework, while being a promising step in addressing the issues of fake news and doxxing, does come with certain limitations that necessitate further research and development. To enhance the framework's accuracy, future work should focus on refining its classification models, possibly through advanced machine learning techniques and more extensive, diverse training datasets. Moreover, the digital landscape is dynamic and ever-evolving, with new tactics and threats continually emerging. As a result, continuous updates and adaptations are essential to make it even more robust and adaptable in the complex online environment.

\bibliography{references}
\end{document}